\documentclass[runningheads]{llncs}
\usepackage[T1]{fontenc}
\usepackage{graphicx}
\usepackage{booktabs}
\usepackage{multirow}
\usepackage{amsmath}
\usepackage{comment}
\usepackage{amssymb}  

\begin{document}
\title{Mapping Similarity Spaces across Embedding Models 
with Synthetic Query Probing}
\titlerunning{Mapping Similarity Spaces across Embedding Models with SQP}
\author{Marcin Rozmus\inst{1}\orcidID{0009-0004-4594-7609} \and
Peter van der Putten\inst{2,3}\orcidID{0000-0002-6507-6896}
}
\authorrunning{M. Rozmus and P. van der Putten}
%

\institute{GenAI Engineering, Pegasystems, Krak\'ow, Poland\\
\email{marcin.rozmus@pega.com}
\and AI Lab, Pegasystems, Amsterdam, The Netherlands\\
\email{peter.van.der.putten@pega.com}
\and LIACS, Leiden University, Leiden, the Netherlands\\
\email{p.w.h.van.der.putten@liacs.leidenuniv.nl}
}%
\maketitle              

\begin{abstract}
Retrieval-Augmented Generation systems rely on similarity scores to retrieve relevant content, yet scores are not directly comparable across embedding models due to differing geometric properties, complicating model migration and limiting threshold reuse. We study how similarity scores can be related by learning mappings between score distributions rather than embeddings. We introduce Synthetic Query Probing, generating queries from documents to create controlled query–chunk pairs, enabling large-scale, reference-free analysis of cross-model similarity behavior. 
We evaluate the approach on multiple embedding configurations and learn score conversion functions using linear, isotonic, and quantile mappings. Experiments on SciFact and a proprietary corpus show that while models largely agree on rankings, their absolute scores exhibit systematic distortions. Learned mappings partially align these spaces and improve threshold portability, with isotonic regression performing best. Our results highlight the need for cross-model calibration and position Synthetic Query Probing as a scalable framework for analyzing embedding comparability.
\keywords{Embedding Models \and  Similarity Calibration}
\end{abstract}

\section{Introduction}

Retrieval-Augmented Generation (RAG) has become a dominant architecture for grounding large language models in domain-specific scientific or enterprise knowledge, either for direct end-user access or as knowledge sources for agents ~\cite{lewis2020rag,gao2024retrievalaugmentedgenerationlargelanguage,plaatetal-agents-2025}. At the core of every RAG pipeline lies vector similarity search: a user query is embedded, compared against a corpus of pre-embedded document chunks, and the most similar chunks are selected as context for generation.

A central assumption in this process is that similarity scores are comparable across embedding models. In practice, however, these scores are produced by models with differing geometric properties, dimensionalities, and training objectives. 
As a result, similarity scores are not directly comparable across models, configurations, or corpora, and thresholds calibrated for one setting do not transfer reliably to another. This complicates model migration in RAG systems, which can happen frequently in practice, and limits our ability to reason systematically about relationships between embedding spaces.
While prior work extensively studies embedding quality and downstream performance, it generally treats similarity scores as model-specific and does not address their comparability across models. Consequently, whether similarity spaces can be systematically related and thresholds transferred remains largely unexplored, despite its practical importance for model migration.

In this work, we investigate how similarity scores from different embedding models can be systematically related, and whether mappings between their similarity spaces can be learned. To our knowledge we are the first to explicitly model and empirically analyze cross-model similarity score distributions and their implications for threshold transfer in RAG systems. We introduce \emph{Synthetic Query Probing (SQP)}, a scalable, reference-free methodology in which queries are automatically generated from document chunks with varying levels of relatedness. This allows us to construct controlled query–document pairs and obtain corresponding similarity scores across multiple embedding models, enabling direct cross-model comparison.
We evaluate SQP on four embedding configurations, analyzing similarity distributions and learning cross-model conversion functions.
This study is making the following contributions:
\begin{enumerate}
\renewcommand{\labelenumi}{(\roman{enumi})}
    \item \textbf{Problem formulation.} We formalize the lack of similarity score comparability across embedding models as an underexplored challenge, limiting our fundamental understanding of embedding spaces and impacting practical concerns such as threshold calibration and model migration.
    \item \textbf{Synthetic Query Probing.} We introduce a scalable, reference-free method for analyzing similarity behavior across embedding models by generating controlled query–document pairs without human annotation.
    \item \textbf{Cross-model similarity analysis.} Using this method, we empirically characterize how similarity distributions differ across models and dimensionalities, showing systematic distortions despite preserved ranking structure.
    \item \textbf{Score calibration and transfer.} We show that these distortions can be modeled using learned transfer functions, and provide practical methods, including isotonic regression and quantile mapping, for converting similarity scores and thresholds across models.
\end{enumerate}


\section{Related Work}

Research on evaluating text embedding models has largely followed two lines: benchmarking downstream task performance and directly analyzing embedding spaces. Neither addresses whether the similarity distributions used to calibrate retrieval thresholds are comparable across models and corpora.

\subsection{Benchmarking Embeddings for Down-Stream Task Performance}

Nowadays, there are very mature benchmarks available that evaluate embeddings across a wide range of tasks, such as BEIR \cite{thakur2021beir} and MTEB \cite{muennighoffetal2023}. As of June 2026, MTEB covers over 1000 languages, 130 tasks and (partial) results for over 400 models (see also MMTEB \cite{enevoldsenetal2025-mmteb}). Embedding models are evaluated in terms of downstream performance for tasks such as clustering, classification, bitext mining, semantic textual similarity (STS), reranking, and retrieval. Benchmark results evolve rapidly and vary substantially across tasks, languages, and domains, motivating both systematic model selection and comparison, and methods for understanding the impact of migration between embedding models.
 
\subsection{Beyond Task Benchmarking: Embedding Space Comparison}

Recent work has shifted attention from aggregate benchmark rankings towards understanding the representational and behavioral properties of embeddings themselves, including representational similarity \cite{kornblith2019similarity,casparietal2024}, retrieval overlap across models \cite{casparietal2024}, and robustness under meaning-preserving perturbations \cite{franketal2026-pteb,frank2026hardertextembeddingbenchmark}.

Caspari et al show that benchmarks alone are a weak guide for selecting embedding models in RAG \cite{casparietal2024}. They quantify model similarity along representational similarity via Centered Kernel Alignment on the embedding geometry, and functional similarity via overlap (Jaccard) and ordering (rank similarity) of retrieved documents. On five BEIR datasets, models cluster mainly by family yet diverge sharply in their top-$k$ results at small $k$, the regime that determines RAG quality, so comparable benchmark scores need not imply comparable retrieval. Their work shows that representational and retrieval behavior, rather than benchmark rank, govern sustainability. We complement this by characterizing the score distributions underlying threshold selection.

A related concern is the stability of embeddings under meaning-preserving variation. The PTEB and HTEB benchmarks~\cite{franketal2026-pteb,frank2026hardertextembeddingbenchmark} show that sentence encoders remain sensitive to paraphrasing and other semantic-preserving transformations. These results motivate our study: If similarity scores vary across both models and input perturbations, raw cosine values and their calibrated thresholds cannot be assumed to transfer across configurations.

Perhaps the closest to our work is the study by Tacheny~\cite{tacheny2026calibratedsimilarityreliablegeometric}. To align embedding similarity scores with human judgments from MTEB-STS \cite{enevoldsenetal2025-mmteb}, rather than performing preprocessing or tuning the embedding model they learn a monotonic calibration for the similarity function using a.o. linear and isotonic regression.

\subsection{Research Gap}

While prior work evaluates task performance, representation similarity, retrieval overlap, or similarity to human judgment, it does not address the transferability of similarity thresholds across models. To the best of our knowledge, no work has systematically compared cosine similarity score distributions across embedding models on multiple corpora and analyzed their implications for threshold transfer.

\section{Methodology: Synthetic Query Probing}

To facilitate practical adoption and reproducibility, we propose a deliberately simple, annotation-free, scalable and configurable methodology and experimental protocol. We model mappings between similarity scores rather than between embeddings themselves, avoiding assumptions about geometric alignment.

Given a corpus $\mathcal{C}$, it is partitioned into a set of chunks $\{c_k\}_{k=1}^N$ using a chosen chunking strategy. A uniformly random subset $\mathcal{C}' \subseteq \mathcal{C}$ is then sampled. For each chunk $c \in \mathcal{C}'$, we generate a set of queries $\mathcal{Q}(c)$ using a large language model, covering varying degrees of semantic relatedness. 
In our experiments, we consider the following query classes:
\begin{itemize}
    \item \textbf{PARAPHRASE} ($\mathcal{Q}_{\text{para}}(c)$): queries that can be answered directly and completely from the chunk $c$.
    \item \textbf{RELEVANT} ($\mathcal{Q}_{\text{rel}}(c)$): queries that are topically related to $c$ but require additional context beyond the chunk.
    \item \textbf{IRRELEVANT} ($\mathcal{Q}_{\text{irr}}(c)$): queries drawn from unrelated domains and exhibiting no semantic connection to $c$.
\end{itemize}
These classes induce controlled relevance labels that enable evaluating score distributions and threshold behavior. Without loss of generality, other synthetic labels of interest could be introduced, as long as the full similarity score range is covered.

We focus on cosine similarity, which despite its limitations, is the dominant choice in production embedding-based retrieval systems \cite{reimers2019sentencebert,karpukhin2020dpr,you2025semanticsanglecosinesimilarity} but the method is open to using other distance metrics that address its limitations \cite{you2025semanticsanglecosinesimilarity}. Let $\mathcal{M} = \{m_i\}_{i=1}^K$ denote a set of embedding models. For each model $m_i \in \mathcal{M}$, we compute similarity scores  $s_{m_i}(q, c) \in \mathbb{R}$ for all query–chunk pairs $(q, c)$ with $q \in \mathcal{Q}(c)$, enabling direct cross-model comparison and subsequent analysis, such as fitting similarity score mapping functions across models.

\section{Experiments and Results}

We evaluate Synthetic Query Probing across two corpora and four configurations, measuring per-class similarity distributions, classification threshold precision, and cross-model conversion accuracy. All experiments follow the same framework, enabling comparison across models.

\subsection{Experimental Setup}

We evaluate our approach on two corpora: an open scientific corpus and a large proprietary enterprise knowledge base. Together they test whether score distributions and conversion functions are corpus-dependent.

The SciFact corpus contains 5,183 life-science and biomedical abstracts \cite{wadden-etal-2020-fact}. It is a homogeneous corpus with structured domain vocabulary, expected to produce cleaner semantic separation. The enterprise corpus comprises 114,648 chunks from 33,492 internal and external Pegasystems documents, including product guides, API references, knowledge-base articles, marketing materials, and community content, yielding broad cross-topic similarity.

To trade off coverage, cost, and stability of distributions we generate synthetic query probing data by sampling 100 chunks from each corpus, and use Claude Sonnet 4.6 to generate 10 questions per chunk for each of the 3 relevance classes outlined in the previous section. This process yields 3,000 question--chunk pairs per corpus with ground-truth relevance labels. For the embedding experiments, we use Amazon Titan Text Embeddings V2 at 256, 512 and 1024 dimensions as well as OpenAI text-embedding-ada-002 with 1,536 dimensions.\setcounter{footnote}{0}\footnote{\url{https://www.anthropic.com/news/claude-sonnet-4-6}, \url{https://openai.com/index/new-and-improved-embedding-model/}, \url{https://docs.aws.amazon.com/ai/responsible-ai/titan-text-embeddings/}} 

These question--chunk pairs, along with similarity scores, allow us to compare various embedding configurations, for instance by constructing `cross-configuration mapping functions'. 
These functions can be used in practice, for instance to map minimum similarity thresholds in retrieval from one model to another, but also provide deeper fundamental insight into the nature and behavior of models.
For each corpus, every question--chunk pair has all four cosine similarity scores, we can treat each pairwise mapping as a paired regression problem. All 12 directional pairs are fitted:
Titan 256\,$\to$\,512, 512\,$\to$\,256,
256\,$\to$\,1024, 1024\,$\to$\,256,
512\,$\to$\,1024, 1024\,$\to$\,512,
256\,$\to$\,Ada, Ada\,$\to$\,256,
512\,$\to$\,Ada, Ada\,$\to$\,512,
1024\,$\to$\,Ada, Ada\,$\to$\,1024.
For each pair we evaluate three conversion approaches, linear and isotonic regression, and quantile (CDF) mapping.

We also leverage the synthetic query probing data to carry out a `threshold analysis'.
For each corpus and each model $m_i \in \mathcal{M}$ independently, we perform a binary classification sweep over the similarity scores $s_{m_i}(q, c)$. The positive class collapses PARAPHRASE and RELEVANT queries, $\mathcal{Q}_{\text{para}}(c) \cup \mathcal{Q}_{\text{rel}}(c)$, and the negative class is $\mathcal{Q}_{\text{irr}}(c)$. A query $q$ is classified as positive at threshold $\tau$ if $s_{m_i}(q, c) \geq \tau$. We sweep $T = 500$ evenly spaced candidates

\begin{equation}
\mathcal{T}_{m_i} = \left\{
 \min_{m_i} + k \cdot \frac{\max_{m_i} - \min_{m_i}}{T - 1}
 \;\middle|\; k = 0, 1, \ldots, T-1
\right\}
\end{equation}

where $\min_{m_i} = \min_{(q,c)} s_{m_i}(q,c)$ is the observed minimum and $\max_{m_i} = \max_{(q,c)} s_{m_i}(q,c)$ is the observed maximum of model $m_i$ on the evaluation set. At each $\tau \in \mathcal{T}_{m_i}$ we compute precision, recall, and F1.

\subsection{Per-Class Similarity Distributions}

Several observations emerge from the per-class distribution analysis (Table~\ref{tab:stats_combined}).
First, the Titan dimensionalities form a consistent gradient: as dimensionality decreases, all class means shift upward. The effect is visible in the enterprise and SciFact corpora.
In contrast, Ada occupies a fundamentally different score space in both corpora. Its similarity scores are compressed into a relatively narrow high range, with standard deviations that are two to six times smaller than those of the Titan variants. As a result, the separation between classes is much smaller: the gap between IRRELEVANT and RELEVANT means, depending on the corpus, is two to four times smaller than for the Titan variants.

\begin{table}[tb]
\centering
\caption{Cosine-similarity statistics — SciFact and Enterprise corpora.}
\label{tab:stats_combined}
\small
\begin{tabular}{@{}lllccccccc@{}}
\toprule
\textbf{Corpus} & \textbf{Model} & \textbf{Class} &
\textbf{Mean} & \textbf{Std} & \textbf{Min} &
\textbf{25\%} & \textbf{Median} & \textbf{75\%} & \textbf{Max} \\
\midrule
\multirow{12}{*}{SciFact}
& \multirow{3}{*}{Titan 256}
& PARAPHRASE & 0.54 & 0.17 &  0.04 & 0.43 & 0.56 & 0.67 & 0.90 \\
& & RELEVANT   & 0.39 & 0.14 &  0.04 & 0.29 & 0.38 & 0.47 & 0.89 \\
& & IRRELEVANT & 0.06 & 0.06 & $-$0.13 & 0.02 & 0.06 & 0.10 & 0.25 \\
\cmidrule(l){2-10}
& \multirow{3}{*}{Titan 512}
& PARAPHRASE & 0.52 & 0.17 &  0.02 & 0.40 & 0.53 & 0.64 & 0.89 \\
& & RELEVANT   & 0.36 & 0.14 &  0.00 & 0.26 & 0.35 & 0.45 & 0.88 \\
& & IRRELEVANT & 0.03 & 0.05 & $-$0.10 & 0.01 & 0.03 & 0.07 & 0.19 \\
\cmidrule(l){2-10}
& \multirow{3}{*}{Titan 1024}
& PARAPHRASE & 0.52 & 0.17 &  0.01 & 0.41 & 0.54 & 0.65 & 0.89 \\
& & RELEVANT   & 0.37 & 0.13 &  0.05 & 0.27 & 0.36 & 0.45 & 0.87 \\
& & IRRELEVANT & 0.02 & 0.04 & $-$0.09 & $-$0.01 & 0.02 & 0.04 & 0.14 \\
\cmidrule(l){2-10}
& \multirow{3}{*}{Ada-002}
& PARAPHRASE & 0.86 & 0.03 &  0.72 & 0.85 & 0.87 & 0.88 & 0.93 \\
& & RELEVANT   & 0.84 & 0.03 &  0.75 & 0.82 & 0.84 & 0.86 & 0.92 \\
& & IRRELEVANT & 0.69 & 0.02 &  0.63 & 0.68 & 0.69 & 0.71 & 0.78 \\
\midrule
\multirow{12}{*}{Enterprise}
& \multirow{3}{*}{Titan 256}
& PARAPHRASE & 0.59 & 0.15 &  0.10 & 0.49 & 0.60 & 0.70 & 0.93 \\
& & RELEVANT   & 0.50 & 0.13 &  0.14 & 0.40 & 0.51 & 0.59 & 0.82 \\
& & IRRELEVANT & 0.23 & 0.14 & $-$0.09 & 0.13 & 0.22 & 0.33 & 0.73 \\
\cmidrule(l){2-10}
& \multirow{3}{*}{Titan 512}
& PARAPHRASE & 0.56 & 0.16 &  0.08 & 0.45 & 0.57 & 0.68 & 0.92 \\
& & RELEVANT   & 0.46 & 0.13 &  0.10 & 0.37 & 0.47 & 0.56 & 0.82 \\
& & IRRELEVANT & 0.19 & 0.13 & $-$0.09 & 0.09 & 0.18 & 0.28 & 0.67 \\
\cmidrule(l){2-10}
& \multirow{3}{*}{Titan 1024}
& PARAPHRASE & 0.56 & 0.16 &  0.05 & 0.44 & 0.56 & 0.68 & 0.89 \\
& & RELEVANT   & 0.46 & 0.13 &  0.10 & 0.37 & 0.46 & 0.55 & 0.82 \\
& & IRRELEVANT & 0.19 & 0.14 & $-$0.05 & 0.08 & 0.18 & 0.27 & 0.72 \\
\cmidrule(l){2-10}
& \multirow{3}{*}{Ada-002}
& PARAPHRASE & 0.85 & 0.03 &  0.74 & 0.83 & 0.86 & 0.88 & 0.96 \\
& & RELEVANT   & 0.83 & 0.03 &  0.72 & 0.80 & 0.83 & 0.85 & 0.93 \\
& & IRRELEVANT & 0.75 & 0.05 &  0.60 & 0.71 & 0.76 & 0.79 & 0.89 \\
\bottomrule
\end{tabular}
\end{table}

\begin{figure}[tb]
\centering
\includegraphics[width=1\textwidth]{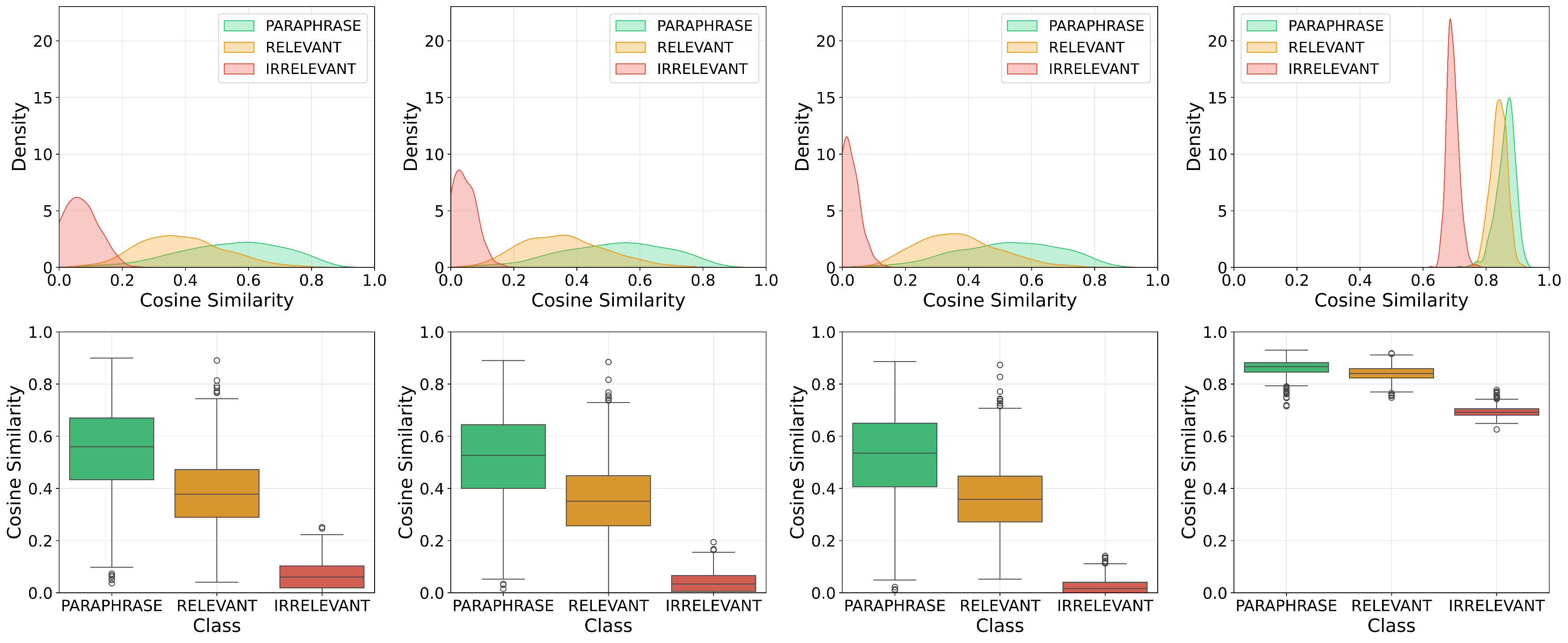}
\caption{Cosine similarity distributions by semantic class for all four
embedding configurations — SciFact corpus. \textit{Top row}: KDE density plots
showing each configuration independently. Titan variants show near-perfect class
separation with IRRELEVANT scores clustered near zero (mean 0.018 at 1024-d,
rising to 0.061 at 256-d), while Ada compresses all classes into a 0.626-0.930
band yet keeps IRRELEVANT fully disjoint from the positive classes.
\textit{Bottom row}: Box-plots confirming the upward IRRELEVANT shift at lower
Titan dimensions and the cleaner class boundaries on SciFact compared to
enterprise content.}
\label{fig:dist_sf_all}
\end{figure}

\begin{figure}[tb]
\centering
\includegraphics[width=1\textwidth]{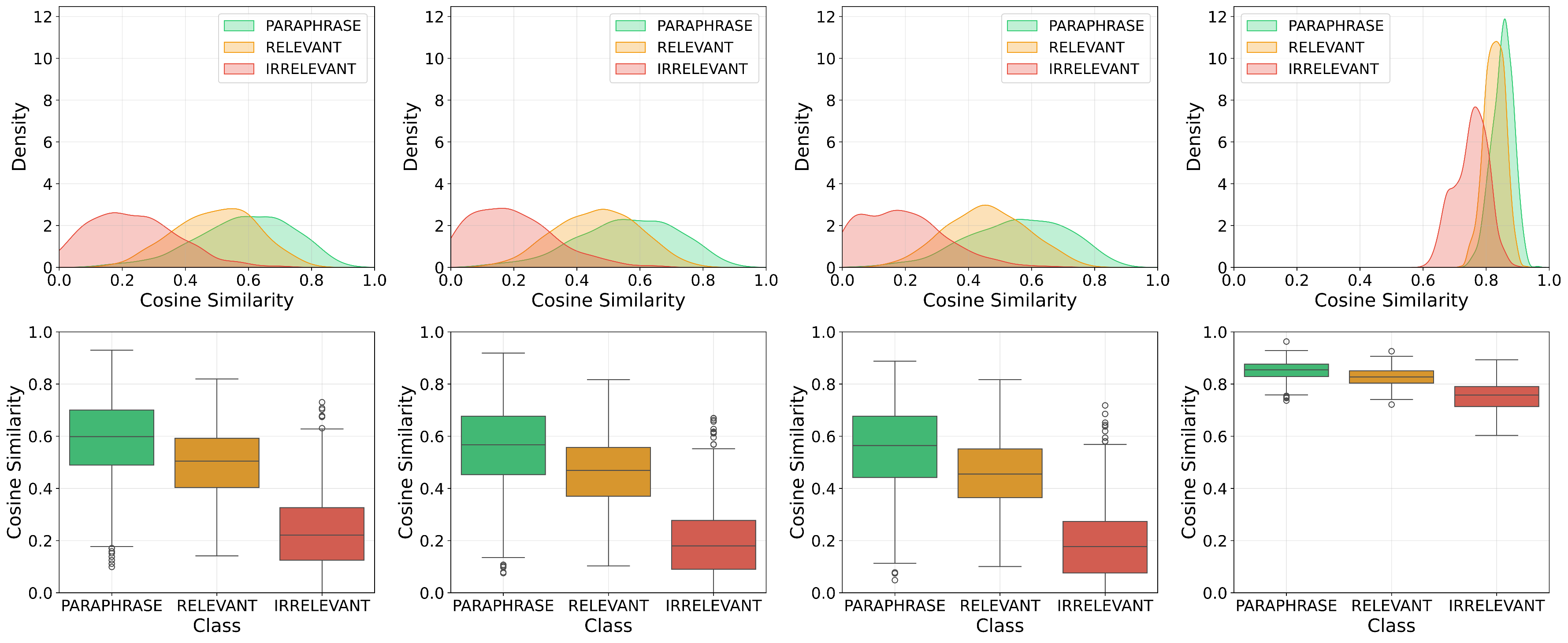}
\caption{Cosine similarity distributions by semantic class for all four
embedding configurations - Enterprise corpus. \textit{Top row}: KDE density
plots showing each configuration independently. Titan variants show clear class
separation with IRRELEVANT scores clustered around 0.2, while Ada compresses
all classes into a narrow 0.603-0.963 band. \textit{Bottom row}: Box-plots
confirming the upward shift in Titan scores at lower dimensions and Ada's much
tighter distributions.}
\label{fig:distributions}
\end{figure}

Despite these differences in scale and dispersion, class ordering is preserved across all configurations. In every case, PARAPHRASE scores highest, followed by RELEVANT and then IRRELEVANT, and this ordering holds consistently across the mean, median, and all quartiles.
However, overlap between classes is most pronounced for Ada on the enterprise corpus, making threshold selection more challenging in that setting. Notably, the maximum IRRELEVANT score for Ada exceeds the PARAPHRASE mean on the enterprise corpus, while on SciFact the Ada IRRELEVANT scores remain separated from the positive classes.
This reinforces the need for careful per-corpus calibration when working in this score space.
Figure~\ref{fig:dist_sf_all} and Figure~\ref{fig:distributions} visualize these distributions as kernel density estimate (KDE) plots and box-plots for all four configurations on each corpus.

\subsection{Precision-First Threshold Analysis}

For each configuration, we identify threshold operating points targeting three
precision levels: $\geq$\,0.93 (balanced), $\geq$\,0.95 (precision-first), and
$\geq$\,0.97 (high-precision). The binary target defines IRRELEVANT as
negative (0) and PARAPHRASE + RELEVANT as positive (1).

The threshold analysis reveals a consistent pattern across both corpora. On SciFact, achieving a precision of at least 0.95 requires thresholds of 0.063, 0.090, and 0.144 for Titan at 1024, 512, and 256 dimensions respectively. On the enterprise corpus the same operating point demands 0.393, 0.400, and 0.439 - roughly six times higher in absolute terms, reflecting the denser overlap between classes in heterogeneous enterprise content. In both cases, naively reusing a 1024-d threshold at lower dimensionalities would lead to a noticeable drop in precision.

Ada consistently requires a much higher absolute threshold than any Titan variant to reach the same precision level. On SciFact, the 0.95 operating point is at 0.718 for Ada compared to 0.063-0.144 for Titan.
On the enterprise corpus the gap is at 0.821 compared to Titan's 0.393-0.439, reflecting the compressed nature of Ada's score space regardless of corpus.

The recall, however, is corpus-dependent. On SciFact, Ada achieves perfect recall (1.000) against Titan's 0.981-0.996, because the clean separation between IRRELEVANT and positive classes in Ada's score space on that corpus makes avoiding false positives trivial. On the enterprise corpus the situation reverses, Titan variants retain higher recall at equivalent precision. Titan's wider class separation could be a more favorable precision-recall trade-off.

\begin{figure}[tb]
\centering
\includegraphics[width=1\textwidth]{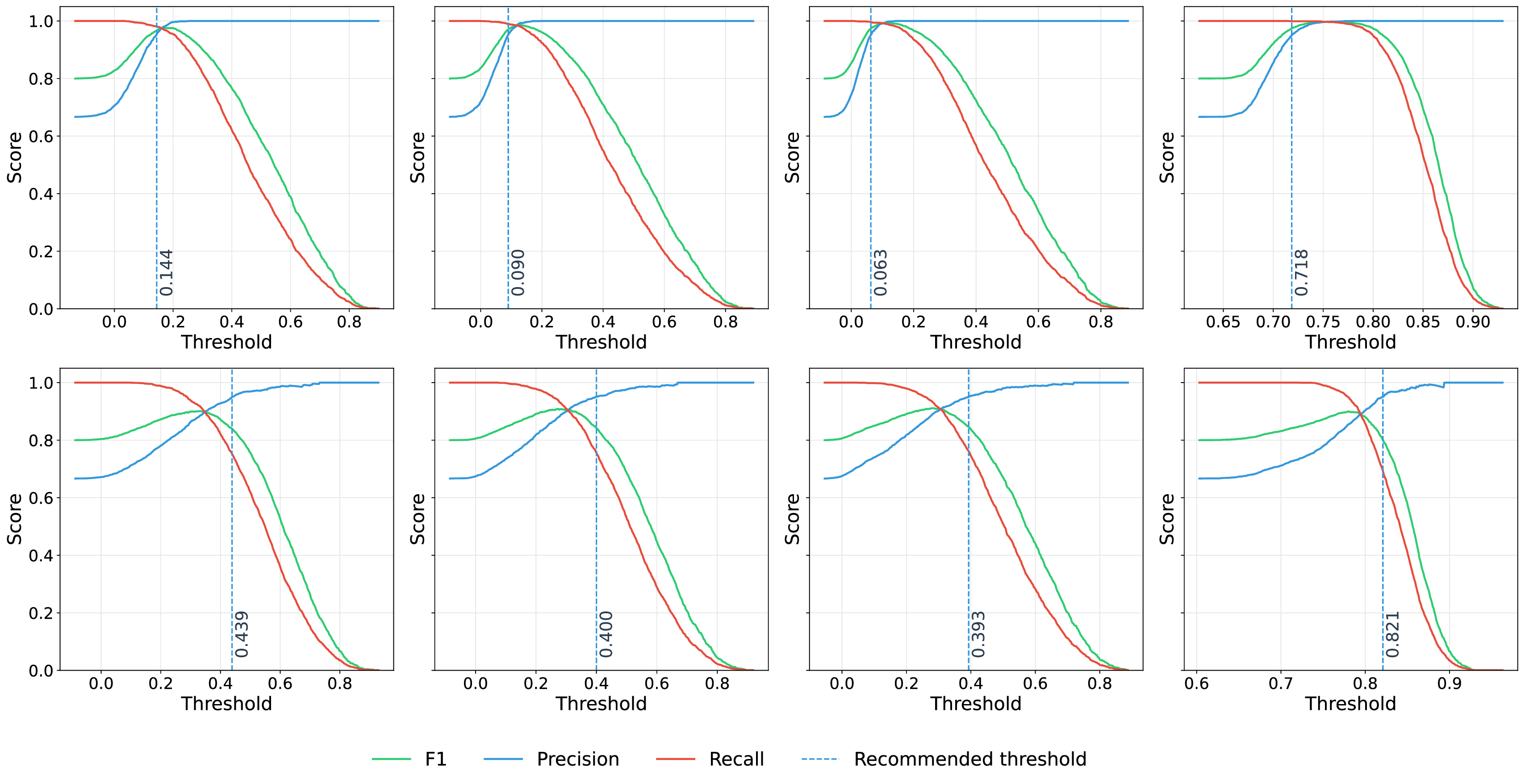}
\caption{Precision, Recall, and F1 versus threshold for all four embedding
configurations on the SciFact and Enterprise corpora. The top row shows SciFact and the bottom row shows Enterprise.}
\label{fig:threshold_curves}
\end{figure}

Despite these differences, false positive counts remain comparable within each corpus. At precision 0.95, all configurations produce 96-104 false positives on SciFact and 72-79 on the enterprise corpus, indicating similar levels of noise rejection even though the underlying threshold scales differ substantially.
Figure~\ref{fig:threshold_curves} shows precision, recall, and F1 as functions of the threshold for all four configurations on each corpus.

\begin{table}[tb]
\centering
\caption{Threshold for the SciFact and Enterprise corpora at precision levels 0.93, 0.95, and 0.97. ``FP'' is the number of IRRELEVANT items above the threshold.}
\label{tab:thresholds_combined}
\small
\begin{tabular}{@{}lllccccc@{}}
\toprule
\textbf{Corpus} & \textbf{Config} & \textbf{Operating Point} &
\textbf{Precision} & \textbf{Threshold} & \textbf{Recall} &
\textbf{F1} & \textbf{FP} \\
\midrule
\multirow{12}{*}{SciFact}
& \multirow{3}{*}{Titan 256}
& Prec $\geq$ 0.93 & 0.93 & 0.128 & 0.985 & 0.957 & 146 \\
& & Prec $\geq$ 0.95 & 0.95 & \textbf{0.144} & 0.981 & 0.966 & 96 \\
& & Prec $\geq$ 0.97 & 0.97 & 0.156 & 0.977 & 0.973 & 59 \\
\cmidrule(l){2-8}
& \multirow{3}{*}{Titan 512}
& Prec $\geq$ 0.93 & 0.93 & 0.080 & 0.993 & 0.960 & 148 \\
& & Prec $\geq$ 0.95 & 0.95 & \textbf{0.090} & 0.990 & 0.969 & 98 \\
& & Prec $\geq$ 0.97 & 0.97 & 0.104 & 0.988 & 0.979 & 56 \\
\cmidrule(l){2-8}
& \multirow{3}{*}{Titan 1024}
& Prec $\geq$ 0.93 & 0.93 & 0.053 & 0.998 & 0.963 & 149 \\
& & Prec $\geq$ 0.95 & 0.95 & \textbf{0.063} & 0.996 & 0.973 & 102 \\
& & Prec $\geq$ 0.97 & 0.97 & 0.076 & 0.994 & 0.982 & 59 \\
\cmidrule(l){2-8}
& \multirow{3}{*}{Ada}
& Prec $\geq$ 0.93 & 0.93 & 0.714 & 1.000 & 0.965 & 143 \\
& & Prec $\geq$ 0.95 & 0.95 & \textbf{0.718} & 1.000 & 0.974 & 104 \\
& & Prec $\geq$ 0.97 & 0.97 & 0.726 & 0.999 & 0.985 & 60 \\
\midrule
\multirow{12}{*}{Enterprise}
& \multirow{3}{*}{Titan 256}
& Prec $\geq$ 0.93 & 0.93 & 0.406 & 0.811 & 0.866 & 122 \\
& & Prec $\geq$ 0.95 & 0.95 & \textbf{0.439} & 0.752 & 0.840 & 79 \\
& & Prec $\geq$ 0.97 & 0.97 & 0.506 & 0.605 & 0.745 & 37 \\
\cmidrule(l){2-8}
& \multirow{3}{*}{Titan 512}
& Prec $\geq$ 0.93 & 0.93 & 0.348 & 0.853 & 0.890 & 128 \\
& & Prec $\geq$ 0.95 & 0.95 & \textbf{0.400} & 0.756 & 0.843 & 78 \\
& & Prec $\geq$ 0.97 & 0.97 & 0.462 & 0.625 & 0.760 & 38 \\
\cmidrule(l){2-8}
& \multirow{3}{*}{Titan 1024}
& Prec $\geq$ 0.93 & 0.93 & 0.348 & 0.846 & 0.886 & 126 \\
& & Prec $\geq$ 0.95 & 0.95 & \textbf{0.393} & 0.762 & 0.846 & 78 \\
& & Prec $\geq$ 0.97 & 0.97 & 0.474 & 0.571 & 0.719 & 34 \\
\cmidrule(l){2-8}
& \multirow{3}{*}{Ada}
& Prec $\geq$ 0.93 & 0.93 & 0.812 & 0.767 & 0.842 & 110 \\
& & Prec $\geq$ 0.95 & 0.95 & \textbf{0.821} & 0.694 & 0.802 & 72 \\
& & Prec $\geq$ 0.97 & 0.97 & 0.840 & 0.521 & 0.678 & 32 \\
\bottomrule
\end{tabular}
\end{table}

\subsection{Cross-Configuration Score Relationships}

\begin{figure}[tb]
\centering
\includegraphics[width=1\textwidth]{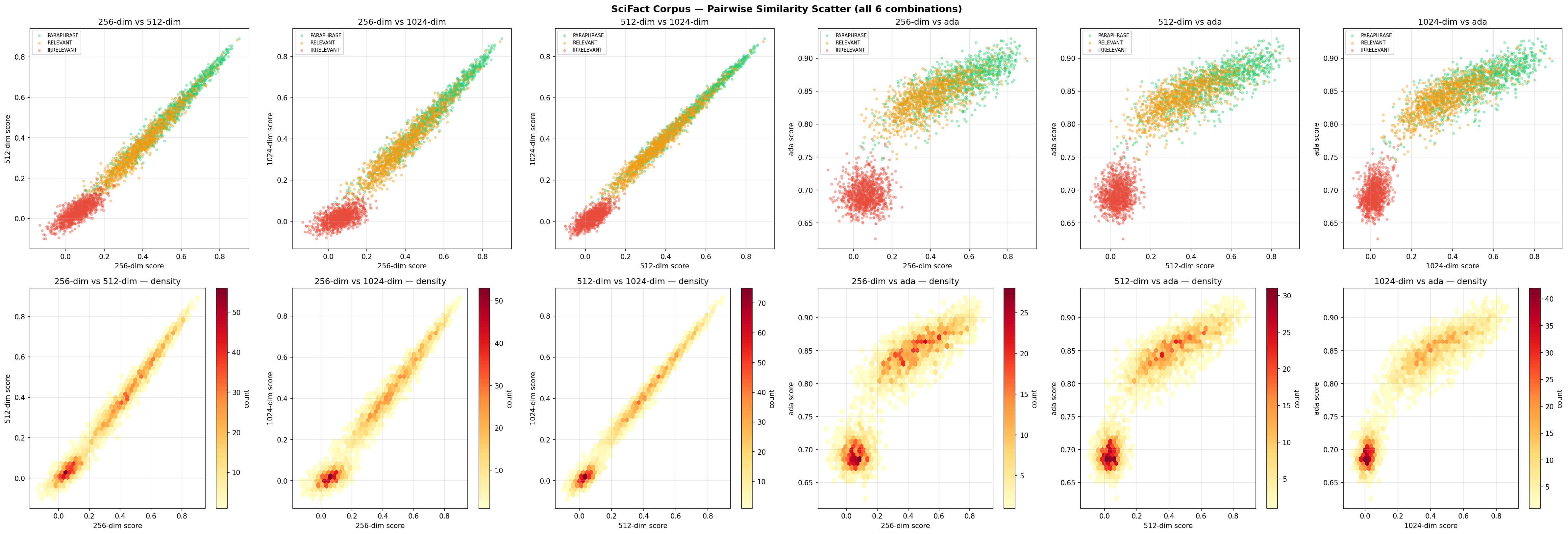}
\caption{SciFact corpus: cross-configuration score relationships for all six pairs. Top row: Scatter plots colored by semantic class---PARAPHRASE (orange) in the upper-right, IRRELEVANT (red) in the lower-left. The Titan cross-dimension pairs show near-linear relationships, reflecting the topical homogeneity of scientific abstracts. The Titan--Ada pairs preserve the S-shaped non-linearity, but IRRELEVANT scores (red) are cleanly isolated in the lower-left with no overlap into PARAPHRASE territory.
Bottom row: Hexagonally binned density plots confirming the sharper cluster boundaries and denser mapping curve on the scientific corpus.}
\label{fig:scifact_scatter}
\end{figure}

\begin{figure}[tb]
\centering
\includegraphics[width=1\textwidth]{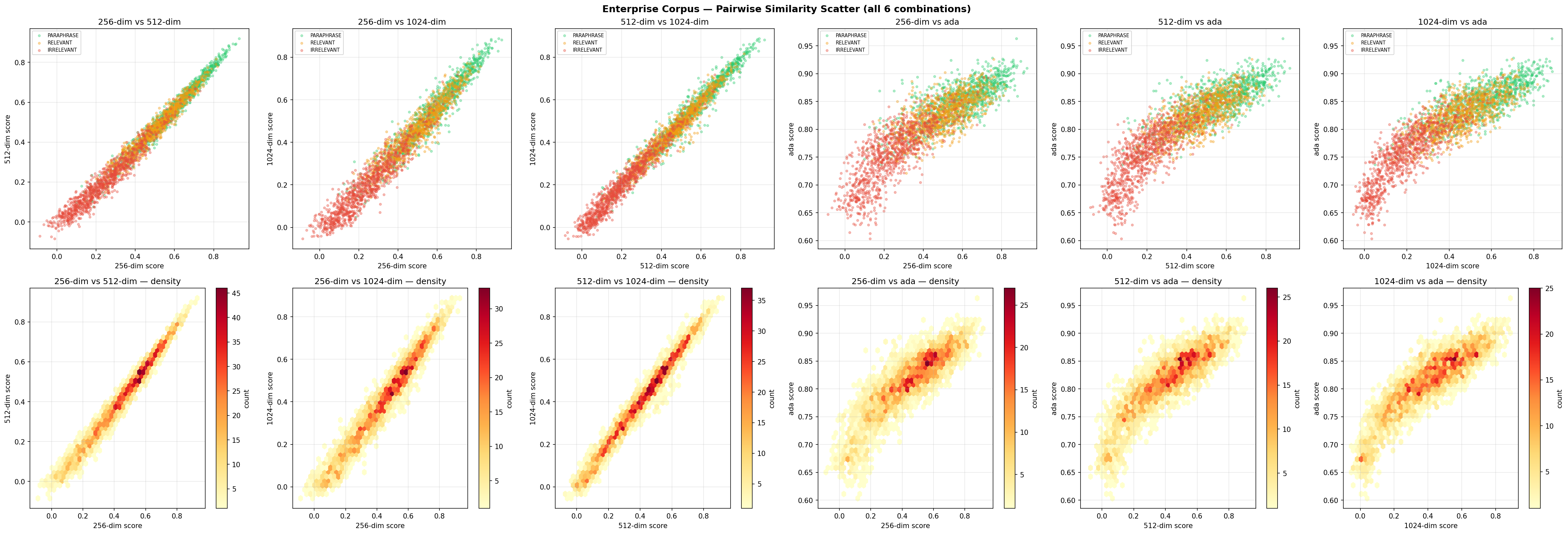}
\caption{Enterprise corpus: cross-configuration score relationships for all six pairs.
Top row: Scatter plots colored by semantic class---PARAPHRASE (orange) in the upper-right, IRRELEVANT (red) in the lower-left. The Titan cross-dimension pairs (256--512, 256--1024, 512--1024) show near-linear relationships with tight scatter. The Titan--Ada pairs exhibit a characteristic S-shaped non-linearity.
Bottom row: Hexagonally binned density plots revealing the concentration of observations along the mapping curve.}
\label{fig:scatter}
\end{figure}

Since every row in the dataset has all four cosine similarity scores for the same question-chunk pair, we can directly visualize the pairwise relationships. Figure~\ref{fig:scifact_scatter} and~\ref{fig:scatter} show scatter and density plots for all six unique configuration pairs.

The scatter analysis reveals several clear patterns. 
First, the Titan cross-dimension pairs (256--512, 256--1024, and 512--1024) exhibit tight, approximately linear relationships with high correlation in both corpora, reflecting that these configurations are simply different dimensional variants of the same underlying model. This linearity is even more pronounced on SciFact, where the topical homogeneity of scientific abstracts reduces pair scatter in comparison to the heterogeneous enterprise corpus. 

The Titan–Ada pairs display an S-shaped non-linearity in both corpora: at lower Ada scores (around 0.65), the corresponding Titan scores are widely dispersed (roughly 0.0-0.4), whereas at higher Ada scores (around 0.90), the Titan scores converge within a narrower band (approximately 0.5-0.8). This behavior provides a clear motivation for using non-linear conversion methods.

Despite these differences in geometry, class separation is consistently preserved across all projections. In SciFact, the IRRELEVANT scores are cleanly isolated in the lower-left of each plot with no overlap into positive class area, confirming that the underlying signal is strongly defined in the scientific corpus.
On the enterprise corpus, the three semantic classes form distinct clusters in every scatter plot, though some overlap is visible between IRRELEVANT and the positive classes in the Titan-Ada projections.

\subsection{Conversion Model Evaluation}

\begin{figure}
    \centering
    \includegraphics[width=1\linewidth]{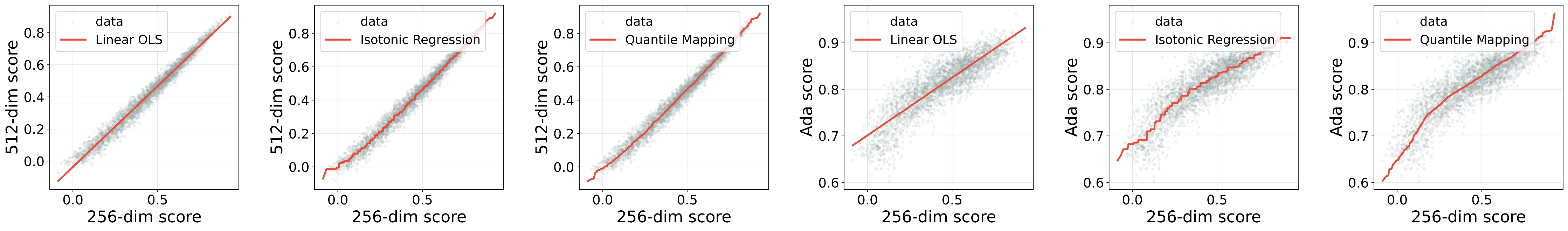}
    \caption{Cross-configuration conversion fits on the SciFact corpus for two representative dimension pairs, each shown with the linear OLS, isotonic regression, and quantile mapping conversion functions overlaid on similarity-score pairs.}
    \label{fig:conversion_fits_selected}
\end{figure}

Table~\ref{tab:conversion} 
compares the three conversion methods across all twelve directional pairs on the SciFact and the enterprise corpora respectively. 
For every directional pair and method we fit the corresponding parameters: the slope and intercept $(a, b)$ for linear OLS, the monotone step function (knot set) for isotonic regression, and the empirical percentile lookup for quantile mapping. To keep this paper readable we omit the full per-pair parameter sets and report only aggregate accuracy (MAE and R\textsuperscript{2}) here. Figure~\ref{fig:conversion_fits_selected} illustrates these fitted conversion functions for two representative directional pairs.

\begin{table}[tb]
\centering
\caption{Conversion model accuracy for all twelve directional pairs on the SciFact and enterprise corpora. Bold indicates the best method for each direction.}
\label{tab:conversion_both}
\scriptsize
\setlength{\tabcolsep}{2pt}

\begin{minipage}[t]{0.49\textwidth}
\centering
\textbf{(a) SciFact corpus}

\vspace{1mm}
\resizebox{\linewidth}{!}{%
\label{tab:conversion_sf}
\begin{tabular}{@{}llccccc@{}}
\toprule
& & \multicolumn{2}{c}{\textbf{Linear}} &
\multicolumn{2}{c}{\textbf{Iso.}} & \textbf{Quant.} \\
\cmidrule(lr){3-4} \cmidrule(lr){5-6} \cmidrule(lr){7-7}
\textbf{From} & \textbf{To} & \textbf{MAE} & \textbf{R\textsuperscript{2}} &
\textbf{MAE} & \textbf{R\textsuperscript{2}} & \textbf{MAE} \\
\midrule
\multicolumn{7}{@{}l}{\textit{Titan $\leftrightarrow$ Titan}} \\
256  & 512  & 0.0265 & 0.980 & \textbf{0.0233} & \textbf{0.984} & 0.0246 \\
512  & 256  & 0.0265 & 0.980 & \textbf{0.0252} & \textbf{0.981} & 0.0267 \\
256  & 1024 & 0.0344 & 0.967 & \textbf{0.0286} & \textbf{0.976} & 0.0302 \\
1024 & 256  & 0.0336 & 0.967 & \textbf{0.0325} & \textbf{0.969} & 0.0351 \\
512  & 1024 & 0.0225 & 0.986 & \textbf{0.0188} & \textbf{0.990} & 0.0200 \\
1024 & 512  & 0.0221 & 0.986 & \textbf{0.0201} & \textbf{0.988} & 0.0215 \\
\midrule
\multicolumn{7}{@{}l}{\textit{Titan $\leftrightarrow$ Ada}} \\
256  & Ada  & 0.0294 & 0.795 & \textbf{0.0177} & \textbf{0.901} & 0.0196 \\
Ada  & 256  & 0.0870 & 0.795 & \textbf{0.0727} & \textbf{0.848} & 0.0814 \\
512  & Ada  & 0.0288 & 0.806 & \textbf{0.0159} & \textbf{0.924} & 0.0179 \\
Ada  & 512  & 0.0833 & 0.806 & \textbf{0.0665} & \textbf{0.865} & 0.0732 \\
1024 & Ada  & 0.0261 & 0.841 & \textbf{0.0143} & \textbf{0.945} & 0.0162 \\
Ada  & 1024 & 0.0769 & 0.841 & \textbf{0.0604} & \textbf{0.889} & 0.0657 \\
\bottomrule
\end{tabular}%
}
\end{minipage}
\hfill
\begin{minipage}[t]{0.49\textwidth}
\centering
\textbf{(b) Enterprise corpus}

\vspace{1mm}
\resizebox{\linewidth}{!}{%
\label{tab:conversion}
\begin{tabular}{@{}llccccc@{}}
\toprule
& & \multicolumn{2}{c}{\textbf{Linear}} &
\multicolumn{2}{c}{\textbf{Iso.}} & \textbf{Quant.} \\
\cmidrule(lr){3-4} \cmidrule(lr){5-6} \cmidrule(lr){7-7}
\textbf{From} & \textbf{To} & \textbf{MAE} & \textbf{R\textsuperscript{2}} &
\textbf{MAE} & \textbf{R\textsuperscript{2}} & \textbf{MAE} \\
\midrule
\multicolumn{7}{@{}l}{\textit{Titan $\leftrightarrow$ Titan}} \\
256  & 512  & 0.0195 & 0.985 & \textbf{0.0176} & \textbf{0.987} & 0.0189 \\
512  & 256  & 0.0194 & 0.985 & \textbf{0.0177} & \textbf{0.987} & 0.0189 \\
256  & 1024 & 0.0228 & 0.980 & \textbf{0.0204} & \textbf{0.983} & 0.0218 \\
1024 & 256  & 0.0223 & 0.980 & \textbf{0.0203} & \textbf{0.983} & 0.0218 \\
512  & 1024 & 0.0107 & 0.995 & \textbf{0.0096} & \textbf{0.996} & 0.0101 \\
1024 & 512  & 0.0108 & 0.995 & \textbf{0.0096} & \textbf{0.996} & 0.0101 \\
\midrule
\multicolumn{7}{@{}l}{\textit{Titan $\leftrightarrow$ Ada}} \\
256  & Ada  & 0.0224 & 0.769 & \textbf{0.0195} & \textbf{0.820} & 0.0209 \\
Ada  & 256  & 0.0814 & 0.769 & \textbf{0.0742} & \textbf{0.820} & 0.0780 \\
512  & Ada  & 0.0222 & 0.778 & \textbf{0.0194} & \textbf{0.826} & 0.0206 \\
Ada  & 512  & 0.0800 & 0.778 & \textbf{0.0733} & \textbf{0.826} & 0.0770 \\
1024 & Ada  & 0.0219 & 0.780 & \textbf{0.0192} & \textbf{0.830} & 0.0204 \\
Ada  & 1024 & 0.0796 & 0.780 & \textbf{0.0730} & \textbf{0.918} & 0.0765 \\
\bottomrule
\end{tabular}%
}
\end{minipage}

\end{table}

Cross-dimension conversions prove to be highly accurate on both corpora. Titan\,$\leftrightarrow$\,Titan mappings reach R\textsuperscript{2} values of at least 0.967 on SciFact corpus (MAE $\leq$ 0.035) and at least 0.980 on the enterprise corpus (MAE $\leq$ 0.023) across all methods. In both cases the 512\,$\leftrightarrow$\,1024 pair is the tightest (R\textsuperscript{2} =  0.990 SciFact, 0.996 enterprise. MAE $\approx$ 0.010-0.019), confirming that these two dimensionalities are nearly identical. 

By contrast, cross-model conversions are inherently noisier, and their accuracy is markedly corpus-dependent. On the heterogeneous enterprise corpus, Titan\,$\leftrightarrow$\,Ada mappings achieve R\textsuperscript{2} values of only 0.820-0.918 with isotonic regression. On the cleaner, topically homogeneous SciFact corpus the same conversions are substantially tighter. The Titan\,$\leftrightarrow$\,Ada direction reaches R\textsuperscript{2} = 0.848-0.945 (e.g.\ 0.945 for 1024\,$\to$\,Ada, MAE = 0.014), and the harder Ada\,$\to$\,Titan direction improves to 0.85-0.89. This gap mirrors the per-class separation reported earlier: when IRRELEVANT queries are cleanly isolated, the cross-model relationship is easier to fit. 
The lower rows of SciFact figure make the difference visible: the characteristic S-shaped non-linearity of the Titan--Ada mapping is captured by isotonic regression but systematically missed by linear OLS, and the SciFact scatter is noticeably denser and better separated than the enterprise scatter. In both corpora these results remain below the cross-dimension figures but stay practically useful. The residual noise reflects the fundamentally different architectures underlying the two models.

Across all evaluated pairs and both corpora, isotonic regression consistently delivers the best performance, achieving the highest R\textsuperscript{2} values alongside the lowest MAE. This demonstrates its advantage for this type of conversion task.

Finally, there is a notable asymmetry in error magnitude: the MAE for Ada\,$\to$\,Titan conversions is roughly three to four times higher than for the reverse Titan\,$\to$\,Ada direction in both corpora (e.g.\ 0.073 vs.\ 0.019 on the enterprise corpus and 0.060 vs.\ 0.014 on SciFact, for the 1024\,$\leftrightarrow$\,Ada pair). This difference arises because Titan operates over a wider score range, which amplifies prediction errors in absolute terms even though the underlying fit quality (R\textsuperscript{2}) is identical in both directions.

\section{Discussion}

Our experiments demonstrate the value of Synthetic Query Probing (SQP) as a reference-free framework for analyzing similarity score behavior across embedding models. By focusing on score distributions and threshold calibration, orthogonal to downstream ranking quality, we show that similarity functions exhibit approximately preserved ranking structure but systematic scale distortion across models and similarity ranges.

A key finding is the existence of two regimes of conversion difficulty. Cross-dimension mappings within the Titan family are near-lossless and largely corpus-invariant (R\textsuperscript{2} $\geq$ 0.97), reflecting their shared underlying representation. In contrast, cross-model mappings (Titan\,$\leftrightarrow$\,Ada) are inherently noisier, non-linear, and corpus-dependent, exhibiting an S-shaped relationship that is best captured by isotonic regression. Conversion accuracy is substantially higher on homogeneous corpora (SciFact) than on heterogeneous enterprise data, indicating that cross-model comparability is not a fixed property of model pairs but depends on corpus characteristics.

More broadly, conversion quality is strongly driven by class separability. When relevant and irrelevant content are clearly separated, similarity mappings are tighter and more predictable; when overlap increases, mapping accuracy degrades, even for the same model pair. This suggests that SQP-derived separation statistics can serve as a prior indicator of expected conversion reliability before fitting any transfer function.

We also observe an asymmetry in the sources of threshold variation. Changing embedding dimensionality within a model family has a relatively small effect on optimal thresholds, whereas changing the corpus induces large shifts. This implies that thresholds are primarily corpus-dependent, and that per-corpus calibration remains necessary even when the embedding model is held fixed. 

Across all experiments, isotonic regression consistently provides the best conversion accuracy. Its ability to model monotonic but non-linear relationships makes it particularly suitable for cross-model calibration, whereas linear mappings suffice only for near-linear regimes. We also observe directional asymmetry: conversions from compressed score spaces (e.g., Ada) to wider ones (e.g., Titan) incur higher absolute error, reflecting differences in scale rather than fit quality.

From a practical perspective, these findings imply that lightweight linear mappings may suffice for intra-family changes (e.g., dimensionality reduction), but cross-model migration requires non-linear calibration and explicit validation on the target corpus. More importantly, threshold reuse across corpora is considerably less reliable than reuse across model configurations.

As the focus of this paper is to prove the method in principle, several limitations still qualify these results. We evaluate only two corpora, and broader validation is required to assess generality, including additional sets of models and model families. Synthetic queries may introduce bias, although prior work suggests such benchmarks are predictive of real retrieval behavior \cite{elburgetal2026}. Our analysis assumes normalized embeddings, static thresholds, and single-stage retrieval without re-ranking. Finally, conversion functions are fitted on the full dataset and should be extended with proper train–test protocols for reusable calibration.

Future work includes extending SQP to newer embedding models and a wider set of corpora, modeling conversion quality directly from corpus statistics and down-stream task performance, using SQP for continuous monitoring of corpus drift to trigger re-calibration, and also comparing different methods and models for question generation with SQP itself. Together, these results position SQP as a practical and scalable tool for understanding and aligning embedding similarity spaces in evolving RAG systems.

\section{Conclusion}

Understanding the behavior and comparability of embedding-based similarity functions is critical for both the scientific study of representation spaces and the reliable deployment of applications that rely on semantic embeddings such as Retrieval-Augmented Generation systems. We have shown that similarity scores produced by embedding models are not directly comparable, despite preserving relative ranking structure, and that their distributions exhibit systematic, model- and corpus-dependent distortions. Using Synthetic Query Probing, we provide a scalable, reference-free approach to characterize these differences and to learn transfer functions that partially align similarity spaces. Our results demonstrate that isotonic regression offers an effective and practical solution for cross-model calibration, highlighting the need for explicit threshold calibration when deploying and migrating RAG systems across models and corpora.

\begin{credits}

\subsubsection{\discintname}
The authors have no competing interests to declare that are relevant to the content of this article.
\end{credits}
%
%
%
 \bibliographystyle{splncs04}
 \bibliography{mybib}
\end{document}